\documentclass{article}
\usepackage{ijcai25}

\usepackage{times}
\usepackage{soul}
\usepackage{url}
\usepackage[hidelinks]{hyperref}
\usepackage[utf8]{inputenc}
\usepackage[small]{caption}
\usepackage{graphicx}
\usepackage{amsmath}
\usepackage{amsthm}
\usepackage{booktabs}
\usepackage{algorithm}
\usepackage{algorithmic}
\usepackage[switch]{lineno}
\usepackage{adjustbox}

\usepackage{comment}
\usepackage{setspace}

\usepackage[dvipsnames,svgnames]{xcolor}
\definecolor{darkblue}{RGB}{0,0,150}
\definecolor{darkred}{RGB}{150,0,0}
\definecolor{darkgreen}{RGB}{0,150,0}
\definecolor{mybgcolor1}{HTML}{006d77}
\definecolor{mybgcolor2}{HTML}{edf6f9}

\usepackage{xspace}

\newcommand{\citep}[1]{\cite{#1}}
\newcommand{\etal}{\textit{et~al.}}

\newcommand{\school}{paradigm}
\newcommand{\School}{Paradigm}

\title{Paradigms of AI Evaluation: Mapping Goals, Methodologies and Culture}

\author{
John Burden$^{1}$\thanks{Joint first authorship.}
\and
Marko Tešić$^{1}$\footnotemark[1]
\and
Lorenzo Pacchiardi$^{1}$\footnotemark[1]
\And
Jos\'e Hern\'andez-Orallo$^{1,2}$\\
\affiliations
$^1$Leverhulme Centre for the Future of Intelligence, University of Cambridge\\
$^2$VRAIN, Universitat Polit\`ecnica de Val\`encia\\
\emails
\{jjb205, mt961, lp666\}@cam.ac.uk,
jorallo@upv.es
}

\begin{document}
\everypar{\looseness=-1}

\maketitle

\begin{abstract}
Research in AI evaluation has grown increasingly complex and multidisciplinary, attracting researchers with diverse backgrounds and objectives. As a result, divergent evaluation \textit{{\school}s} have emerged, often developing in isolation, adopting conflicting terminologies, and overlooking each other's contributions. This fragmentation has led to insular research trajectories and communication barriers both among different {\school}s and with the general public, contributing to unmet expectations for deployed AI systems. To help bridge this insularity, in this paper we survey recent work in the AI evaluation landscape and 
identify six main {\school}s. We characterise major recent contributions within each {\school} across key dimensions related to their goals, methodologies and research cultures. By clarifying the unique combination of questions and approaches associated with each \school, 
we aim to increase awareness of the breadth of current evaluation approaches and foster cross-pollination between different {\school}s. We also identify potential gaps in the field 
to inspire future research directions.
\end{abstract}

\section{Introduction}

In recent years, Artificial Intelligence (AI) has advanced rapidly and gained public prominence. In particular, general-purpose AI systems, such as Large Language Models (LLMs), have become widely deployed for real-world applications across various sectors \citep{llm_applications}. 
As AI adoption grows, so does the need to understand AI systems capabilities \citep{hernandez2017measure} and the risks they pose \citep{hendrycks2021unsolved} to ensure responsible deployment.

\looseness=-1
At the same time, the technical expertise required to effectively use state-of-the-art AI models has decreased,  enabling a broader range of researchers and practitioners to engage in AI evaluation. This has brought new perspectives to the field, with evaluators from 
a multitude of disciplines beyond AI, including  
Cognitive Science, Psychology, Economics, Social Sciences, Software and Safety Engineering contributing to the diversity of methodologies and insights.

\begin{figure}[t!]
    \centering \includegraphics[width=1.0\columnwidth]{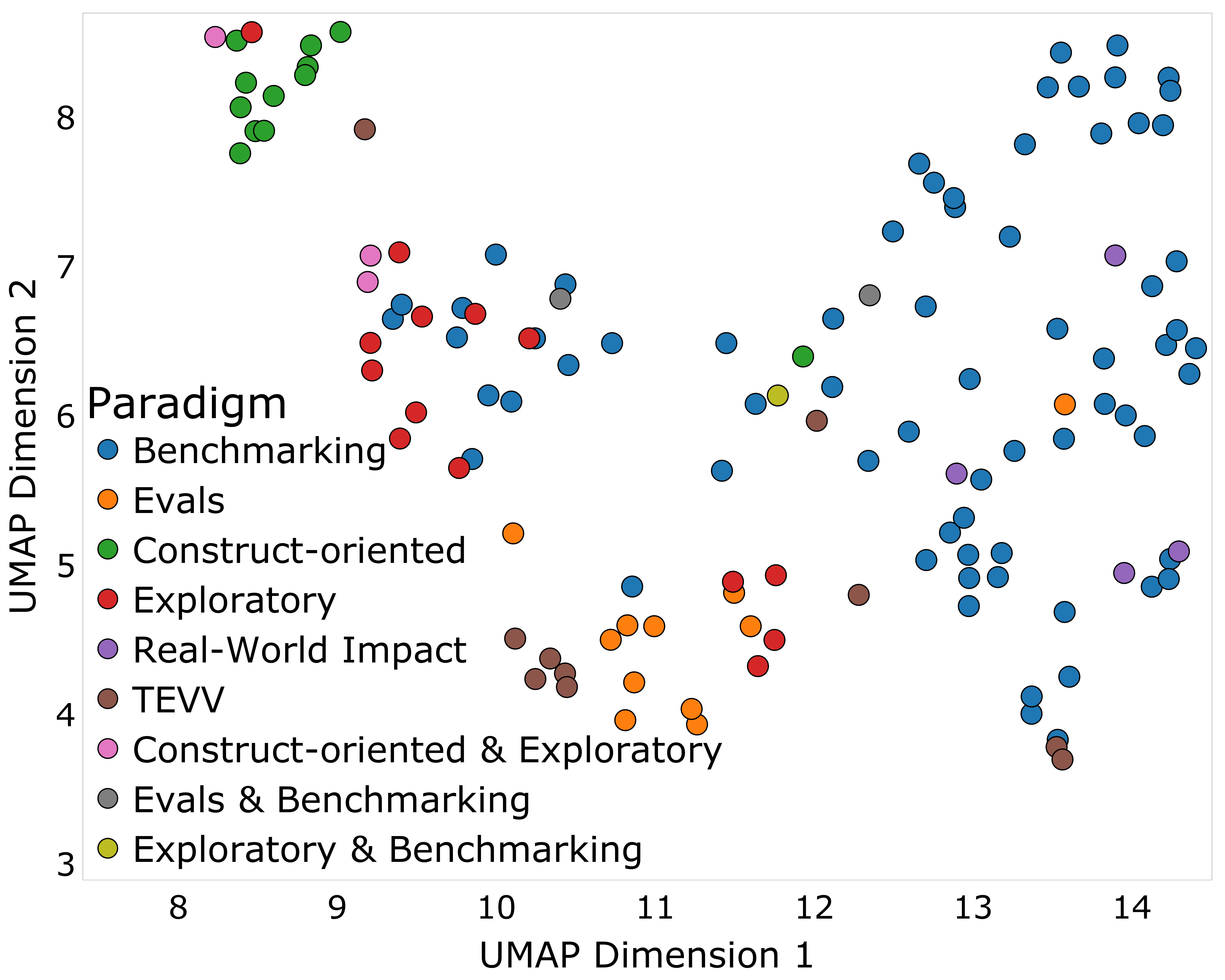}
    \caption{Two-dimensional projection of the surveyed papers based on our framework's dimensions. The coordinates were obtained using UMAP projection \protect\cite{mcinnes2018umap} of the Jaccard-distance matrix. Each point represents a surveyed paper, whose colour indicates the paradigms it belongs to. We find clusters of papers corresponding to different paradigms. The interactive plot and the code to reproduce it are available \protect\href{https://github.com/Kinds-of-Intelligence-CFI/Paradigms-of-AI-Evaluation}{\protect\underline{here}}.}
    \label{fig:main}
\end{figure}

While the influx of varied expertise is essential to evaluating the complex and multifaceted impact of modern AI systems on society, it has created a fragmented AI evaluation landscape,  marked by a lack of standardisation and limited cross-collaboration. As a result,  research efforts have become insular and fail to incorporate advancements from other communities. Communication challenges arise, with key terms often conveying distinct connotations in different communities. For instance, AI developers and regulators rely on evaluations to determine if a system is safe to deploy \citep{phuong2024evaluatingfrontiermodelsdangerous}. In contrast, AI adopters are more concerned with understanding if a system can automate specific tasks within their organisation \citep{asgari2024testing}. Both communities, however, refer to ``capability evaluations''.

Beyond scientific research and industry practices, under-appreciating the breadth of AI evaluation may impact policy, as various policy initiatives ubiquitously rely on ``evaluation". For instance, the EU AI Act states that developers of a general-purpose AI model with ``high impact capabilities evaluated on the basis of appropriate technical tools and methodologies'' \cite[Art. 51(1)]{euaiact} are subject to additional requirements, such as ``perform[ing] model evaluation [...] with a view to identifying and mitigating systemic risks'' \cite[Art. 55(1)]{euaiact}; thus, ``evaluation''  refers, in the same document, both to measurement of capabilities and identification of risks. Ongoing efforts\footnote{Such as the Code of Practice for general-purpose AI models, \url{https://digital-strategy.ec.europa.eu/en/policies/ai-code-practice}.} identifying concrete evaluation tools to satisfy those requirements should consider the full spectrum of AI evaluation practices and understand their terminological and methodological differences, to select the appropriate set of techniques for distinct regulatory purposes.
 
To address this, in this paper we survey the landscape of AI evaluation by collecting 125+ papers representative of the various existing approaches\footnote{We aimed to provide a representative view of AI evaluation rather than an exhaustive catalogue. Hence, we focused on (1) maintaining diversity in the selection of research, (2) emphasising work from the last five years. We have, however, included certain earlier works that have stood the test of time and are still widely used to evaluate modern AI systems. This balance allows us to accurately survey the current landscape of AI evaluation praxis.
} and identify six main \textit{{\school}s}. We define an AI evaluation {\school} as a conceptual framework that groups together studies with similar methodologies, evaluation goals and underlying assumptions that influence the collection of evidence. These {\school}s are not rigid, universally agreed-upon categories but rather our attempt to map and structure the AI evaluation landscape. By delineating them, we aim to provide a useful perspective for researcher and practitioners, helping them navigate and critically assess different evaluation approaches.

To operationalise this framework, we systematically annotated the collected papers according to a series of dimensions of analysis related to goals, methodologies and culture\footnote{We are unable to reference all the surveyed papers within the allocated page limit; the complete annotated list is available \href{https://lorenzo-pacchiardi.notion.site/Paradigms-of-AI-evaluation-1975586880fe80978007cda094edce30}{\underline{here}}.}. This annotation allows us to clarify
the questions and approaches adopted by each paradigm, as well as how they differ from one another. 
Our contributions are twofold: (1) we provide researchers with a broad overview of existing AI evaluation practices and identify main AI evaluation {\school}s; (2) by highlighting differences between {\school}s, we facilitate comparison, collaboration and knowledge exchange across research communities.

\looseness=-1
The paper is organised as follows: Sec.~\ref{sec:dimensions} introduces the dimensions of analysis, while Sec.~\ref{sec:paradigms} describes the identified paradigms. Sec.~\ref{sec:findings} discusses the role these paradigms play within the AI evaluation ecosystem. In  Sec.~\ref{sec:new_works}, we identify gaps in the AI evaluation landscape and explore ways to leverage existing methodologies to advance the field.
Finally, Secs.~\ref{sec:Lims} and \ref{sec:conclusion} discuss the limitations of our survey and present our conclusions.

\subsection{Scope}

For our analysis, we define AI evaluation as {\em the process of  measuring and anticipating the behavioural properties of AI systems and their societal impact to inform decisions about their use}. 
In our survey, we consider work on the evaluation of any kind of AI system, component or algorithm. However, we interpret our definition as excluding explainability and mechanistic interpretability, which aim to get an understanding of the inner workings of AI systems. While such an understanding  may be used to anticipate its behaviour \citep{casper_black-box_2024}, we decide to concentrate on work directly measuring the behavioural properties. We refer interested readers to existing high-quality surveys on interpretability and explainability \citep{yang_survey_2023,bereska_mechanistic_2024}. Moreover, we do not consider purely theoretical papers presenting conceptual frameworks for evaluation \citep{bengio2024can}, methodologies to reduce the computational cost of existing evaluations \citep{kaplan2020scaling,polo2024tinybenchmarks}
and work studying how humans react or think about AI systems \citep{steyvers2025large}.

\subsection{Previous Analyses of AI Evaluation}
Several recent surveys have explored AI evaluation from specific perspectives.
Many of these are focused on LLMs, overviewing  benchmarking \citep{chang2024survey} or red-teaming \citep{lin2025against}, studying how ethical aspects are evaluated \citep{lyu2025ethical} or analysing the literature under a verification and validation perspective \citep{huang2024verification}. Other surveys include Ruah~\textit{et al.,}~\shortcite{rauh2024gaps}, which focuses on the safety of generative AI systems, and \shortcite{gollner2023bridging}, which considers responsible AI principles. In contrast, our work focuses on a wide variety of AI systems and evaluation approaches.

Other work discussed the quality of evaluation practice and instruments \citep{hernandez2020ai,burden2024evaluating}, for instance touching on reproducibility \citep{burnell2023rethink}, statistical rigour \citep{gorman-bedrick-2019-need}, validity \citep{Subramonian2023ItTT} and representativeness and fairness \citep{bergman2023representation,gollner2023bridging}. We do not focus on these issues here, as they apply broadly to all AI evaluation tools.

Bieger~\etal~\shortcite{bieger2016evaluation} proposed a normative framework for evaluating adaptive general-purpose AI, outlining the purposes of evaluation, the properties to be measured, and the challenges involved. In contrast,  we take a descriptive approach to AI evaluation, mapping the landscape by identifying the advantages and limitations of different methodologies without prescribing any particular approach. More recently, Cohn~\etal~\shortcite{cohn2023framework} defined different ``facets'' of evaluation instruments. While there is some overlap with our dimensions of analysis, several of their facets refer to validity and consistency, properties that are broadly relevant across all {\school}s and are hence excluded from our analysis.
While Cohn~\etal~\shortcite{cohn2023framework} apply their framework to 23 evaluation works, neither Bieger~\etal~\shortcite{bieger2016evaluation} nor Cohn~\etal~\shortcite{cohn2023framework} identify distinct {\school}s of evaluation or characterise their defining features. Further, these works do not survey the larger and rapidly evolving AI evaluation landscape shaped by recent advancements in AI.

\section{Dimensions of Analysis}
\label{sec:dimensions}

We explore {\em goals}, {\em methodologies} and {\em cultures} in AI evaluation. These three factors shape the way evaluations are designed, applied, and interpreted, allowing us to highlight the diversity of approaches, clarify underlying assumptions, and identify gaps across different evaluation {\school}s. 

We break down these three factors into key dimensions (each with a discrete set of possible values) that we use to map the landscape of AI evaluation. 
In some cases, a single work may be assigned multiple values for the same dimension.
For example, an evaluation may assess both performance and safety (two distinct values within the Indicator dimension) using the same tasks. Fig.~\ref{fig:main} shows the clustering of surveyed papers based on these dimensions and highlights the relationship between these clusters and assigned paradigms.

\subsection{The Goals}
Evaluation can be marked by the type of insight sought. We make use of the following dimensions:

\noindent  {\bf Indicator (performance, fairness,  safety, robustness and reliability, behavioural features, cost)}. One possible goal of AI evaluation is to determine the {\em performance} of an AI system, namely, its ability to successfully complete tasks. Another important consideration is {\em fairness}: the extent to which demographic groups are treated differently by the AI system. Alternatively, {\em safety} encompasses concerns such as preventing the generation of manipulative or toxic content, mitigating harmful outcomes whether explicitly prompted by a malicious actor or arising from the system’s design and behaviour (alignment). Other evaluation tools focus on the {\em robustness and reliability} of an AI system,  that is, the extent to which its behaviour is affected by factors that are unrelated to the task at hand and how the system fails in presence of anomalies and edge cases. Some evaluations analyse other {\em behavioural features} (such as preferences, tendencies, reasoning patterns, etc.) of models in response to inputs. 
Finally,  the {\em cost} of using a system (whether monetary, environmental, or ethical) can also be a primary indicator used to evaluate a system.

\noindent  {\bf Distribution summarisation (aggregate, extreme, functional, manual inspection)}. Ideally, AI evaluation would fully describe the distribution of subject behaviour conditioned on all possible input values. Yet, in practice, many approaches are limited to reporting summary statistics. Most typical are {\em aggregate} metrics over a set of inputs  (e.g., mean accuracy in benchmarks). 
Alternatively, the {\em extremes} of the distribution over a set of inputs are sometimes reported. This includes both worst-case (e.g., the possibility of accidents) and best-case scenarios (e.g., best performance over a set of prompts). More refined is a {\em functional} description mapping variations in the input (such as task difficulty) to statistics of the conditional distribution of behaviour. This can allow for anticipation and explanation of behaviour (e.g., representing performance as a function of task difficulty). In contrast, a few works instead perform a {\em manual inspection} of the system's behaviour (e.g., describing performance failures in a qualitative manner) rather than summarising the distribution. 

\noindent{\bf Subject (system, component, algorithm)}. AI evaluation often focuses on self-contained {\em systems} that can function without additional components or adaptations   
(e.g., ChatGPT, a planner or a translator). However, the subject of an evaluation study can also be individual {\em components}
designed to be integrated in other systems, often requiring specialised interfaces (such as specific non-linearity functions, the computation of embeddings, or a SAT solver) or {\em algorithms} expressed in programming languages or pseudocode (for instance, Stochastic Gradient Descent and Naive Bayes). Both components and algorithms are typically assessed based on their impact across often multiple systems that employ them.

\subsection{The Methodologies}

A methodology is a collection of practices, principles, and guidelines used to conduct an evaluation. We consider the following dimensions as key factors for categorising methodologies:

\noindent {\bf Measurement (observations, constructs)}. Evaluations can report direct {\em observations} from measurement instruments (e.g., `System X achieved a score of $x\%$ on a benchmark') or explicitly model latent {\em constructs}---underlying factors that explain an AI system's observed behaviour (e.g.,`System X demonstrates low arithmetic capability').

\noindent {\bf Task origin (operation, sample, design)}.
AI systems can be evaluated in real-world {\em operation} (e.g., testing an autonomous vehicle on public roads or assessing whether an LLM improves productivity in a study involving human workers). Alternatively, evaluation can be conducted using a {\em sample} of tasks drawn from a distribution representative of real-world usage (e.g., a sample of journeys). In some cases, tasks may be created by {\em design} (e.g., on a circuit used as a testbed or simulated environment), either because it is difficult, unsafe or unethical to perform evaluation in real-world cases, or because synthetic tasks are thought to enable a more accurate measurement of the desired property.

\noindent {\bf Protocol (fixed, procedural generation, adaptive, interactive)}. The evaluation process may unfold in different ways. A common approach involves testing an AI system on a {\em fixed} set of tasks, defined by the initial input provided to the AI system (even if the interaction between subject and evaluation unfolds differently due to either stochastic evaluation environment or an agent's decisions). Some evaluations use {\em procedural generation} to create new tasks at test-time for each tested system, following a predefined distribution. In other cases, the generation of tasks or the choice amongst a fixed set of instances is {\em adaptive}, chosen based on previous behaviour of the system on other tasks (e.g., adjusting the difficulty of questions in response to its answers to previous questions). Finally, evaluations can be {\em interactive}, where humans guide the process in real-time, probing the system's behaviour through sequential interactions.   

\noindent {\bf Reference (objective, rubric, subjective, no reference)}.  
In some cases, the AI system's outputs are compared to an {\em objective} value, such as a gold standard correct answer. In other cases, a {\em rubric} is applied, either by a human or an AI scorer, to assess outputs that cannot be objectively compared to a determined value.
Alternatively, evaluation may rely on {\em subjective} feedback or judgement (preferences, moral values, etc.) provided by human users at test time or encoded into an automated system (such as a reward model). 
In some cases, there is {\em no reference} answer for comparison.

\noindent  {\bf Task mode (identification, generation)}.
The evaluation may require the subject to {\em identify} an answer from a pre-determined set of options (e.g., selecting a multiple-choice answer or a class label) or to {\em generate} a novel output (e.g., a numerical value in a continuous range or free-form text).

\subsection{The Cultures}
By cultures, we refer to the people involved in the evaluation process, their norms, interests and terminology. We categorise evaluation cultures with the following dimensions:

\noindent {\bf 
Evaluators (researchers, deployers,
regulators)}. 
Different people may create and conduct an evaluation. We define {\em researchers} as those motivated by  understanding and improving AI systems from a scientific perspective, focusing on fundamental insights and advancements. In contrast, {\em deployers} are primarily concerned with commercial viability, customer satisfaction and competitive advantage; they want to determine whether a system is suitable for deployment, assess pricing strategies, ensure safety and enhance brand reputation. Finally, {\em regulators} 
seek guarantees and information on safety, ethics, risks to society, and legal compliance.  
Our definition considers employees of private companies or regulators conducting research for scientific understanding as researchers.

\noindent {\bf Motivation (comparison, understanding, assurance)}. 
Motivation determines focal points for the evaluation, how the results are interpreted, and the suite of techniques it leverages. System {\em comparison} aims to determine the most suitable subject from a set of candidates for a given scenario, assess whether systems can replace or assist humans, and track the progress of successive  generations. 
Another reason for the development of an evaluation tool is seeking {\em understanding} of what causes AI systems to behave in the way they do. This could be via learning to recognise the cognitive processes the AI systems exhibits, or identifying what aspects of a specific input lead to certain types of behaviour. Finally, evaluation may aim to provide {\em assurance}, determining the conditions of reliable operation or a bound on undesirable behaviour, either using empirical methods or formal proofs.

\noindent {\bf Discipline (AI, Psychology, Security, Economics, etc.)}. 
The people developing and conducting an evaluation may belong to different disciplines, such as {\em AI}, {\em psychology}, {\em security}, {\em economics} or others. Their different cultural perspectives and scholarly norms may reflect differences in the methods employed for the evaluation.

\section{Evaluation {\School}s}
\label{sec:paradigms}
Based on our survey, we identified six main {\school}s of AI evaluation. In the following subsections we describe each {\school}, leveraging  insights from our annotation exercise to operationalise the distinctions between them. In Table~\ref{tab:main}, we present the  most common values for each dimension across {\school}s. Where possible, we have chosen names for these paradigms in line with community convention.

\begin{table*}[ht] 
\centering
\begin{adjustbox}{max width=\textwidth} %
\begin{tabular}{p{2.5cm}ccccccc} \toprule
  \textbf{Dimension} &
  \textbf{Benchmarking} &
  \textbf{Evals} &
  \textbf{Construct-Oriented} &
  \textbf{Exploratory} &
  \textbf{Real-World Impact} &
  \textbf{TEVV}
  \\ \midrule 
  {\em Archetype}  & \citep{Imagenet} & \citep{ganguli2022red} & \citep{GauthierAutomated2024} & \citep{berglund2024the} & \citep{collins2023evaluatinglanguagemodelsmathematics} & \citep{Yang2023Safe} \\ 
  \midrule
  {\bf Indicators}      & Performance & \textbf{Safety}/Fairness & - & Performance/Behaviour & Cost/Fairness/\textbf{Performance} & Safety/robustness and reliability  \\ 
  {\bf Dist. summ.}     & \textbf{Aggregate} & Extreme/Aggregate & {\bf Functional} & Aggregate/Manual Inspection & \textbf{Aggregate} & Extreme  \\ 
 
  {\bf Subject} & - & \textbf{System} & \textbf{System} & System & \textbf{System} & \textbf{System} \\
  \midrule
  {\bf Measurement}      & \textbf{Observed} & {\bf Observed} & {\bf Latent} & {\bf Observed} & Observed & Observed  \\ 
  {\bf Task origin}      & - & \textbf{Design} & Design & Design & - & Design/Operation \\ 
  {\bf Protocol}  & Fixed & Interactive & - & - & - & - \\ 
  {\bf Reference} & Objective & Subjective & - & - & Subjective/Rubric & Objective \\
  {\bf Task Mode} & - & \textbf{Generation} & - & - & \textbf{Generation} & - \\
  \midrule
  {\bf Evaluators}      & Researchers & Researchers, Deployers & \textbf{Researchers} & Researchers & - & -  \\ 
  {\bf Motivation} & Comparison & Assurance/Understanding & Understanding & \textbf{Understanding} & \textbf{Comparison} & \textbf{Assurance} \\
  {\bf Disciplines}      & - & Security, Bio & Cognitive Science & - & Social Sciences & Control Theory  \\     

\midrule
\textit{Raw Number} & $72$ & $13$ & $15$ & $18$ & $4$ & $10$ \\
\textit{Percentage} & $57\%$ & $10\%$  & $12\%$ & $14\%$ & $3.2\%$ & $7.9\%$ \\
\bottomrule  
  
\end{tabular}

\end{adjustbox}
\caption{For each {\school} we identify, the table shows an archetypal paper and the values of different dimensions that mostly characterise that {\school}. Bolded entries indicate values that were present across the vast majority of the considered papers within that {\school}. Empty entries indicate that a dimension was not informative for that {\school}. Also included are the number and percentage of papers identified as belonging to each {\school} (note that a few papers bridge multiple {\school}s and are therefore double-counted above).}
\label{tab:main}
\end{table*}

\subsection{Benchmarking {\School}}
Benchmarking can be traced back to the Common Task Framework of the 1980s \citep{koch2024protoscienceepistemicmonoculturebenchmarking}.
A key tenet of benchmarking is that by providing constant test conditions, any variation in a system's responses compared to other systems can be attributed to its intrinsic characteristics. 
Benchmarking, then, involves evaluating AI systems by testing them on standardised sets---or distributions---of instances and summarising and comparing their performance using different {\em aggregate} measures. 
The focus is typically on evaluating the \textit{observed performance} of a system or component (although benchmarks also frequently test for fairness, safety and robustness). Benchmarks are often marked against an \textit{objective} reference, with \textit{identification} (e.g., multiple-choice questions) being a common task mode.
They are widely applied across a wide range of AI systems, including LLMs, RL agents, image classifiers, and more. Benchmarks are inherently built in order to \textit{compare} systems, providing a tool for selecting the most appropriate AI systems or tracking progress. 
Archetypal examples in the Benchmarking {\School} include  ImageNet \citep{Imagenet}, the Arcade Learning Environment \citep{bellemare2013arcade} and MMLU \citep{hendryckstest2021}. Benchmarks also arise in the form of competitions (e.g., the annual RoboCup competition \citep{Kitano1997RoboCupTR}) where entrants compete to demonstrate competence at a particular task.

Benchmarking offers several inherent advantages. First, the use of a standardised test provides a level playing field for all systems being evaluated, while the use of well-defined, objective metrics reduces the ambiguity and allows easy tracking of performance indicators over time. Benchmarks are often publicly available, increasing the transparency of the evaluation process; this can be further improved by reporting the results at the instance level \citep{burnell2023rethink}. However, the benchmarking {\school} also has notable limitations. Open sharing of benchmarks can lead to data contamination \citep{zhou2023dontmakellmevaluation}, with test data being used for training. Relatedly, repeated testing on the same benchmark may induce overfitting, optimising for specific tests rather than generalised to broader tasks \citep{Fang2023Does}. Further, benchmark results are inherently tied to a particular distribution of test items, limiting the generalisability of the measurement \citep{Raji2021AI}. Further discussions on the limitations of benchmarking can be found in Liao~\etal~\shortcite{liao2021are}.

\subsection{Evals {\School} }
The Evals  {\School} focuses on system \textit{safety}, often operationalised as a system's failure to comply with safety specifications or its tendency to exhibit harmful behaviour during tasks that carry risks (e.g., the extent to which it demonstrates so-called ``dangerous capabilities'' \citep{shevlane2023model,phuong2024evaluatingfrontiermodelsdangerous}), although \textit{fairness} is sometimes considered too. This {\School} often employs \textit{extremes} analysis, identifying specific (worst) cases. The aim is providing {\em assurance} on a system's safety or gaining {\em understanding} on what causes the unsafe behaviour. 
A common methodology within the Evals {\School} is ``{red-teaming}'', an \textit{interactive} and adversarial process where humans or automated agents attempt to ``break'' or provoke the system to \textit{generate} undesirable responses. Red-teaming is commonly employed by model \textit{deployers} (e.g., OpenAI \citep{openai_openai_2023}, Anthropic \citep{anthropic_frontier_2023}).
Generally, tasks are {\em designed} in an adversarial way. 
Evals are predominantly applied to general-purpose AI systems such as LLMs, for which assessing potential risks is particularly relevant due to their broad (and unpredictable) range of applications. Since red-teaming is often conducted by human evaluators, there is typically no objective reference against which the system's responses are compared. Instead, evaluation relies on \textit{subjective} human judgment. Archetypal papers from the Evals {\school} are Ganguli~\etal~\shortcite{ganguli2022red}'s work on red-teaming or Kinniment~\etal~\shortcite{kinniment2024evaluatinglanguagemodelagentsrealistic}'s examination of LLM's ability to self-replicate.

Evals offer a systematic way to identify flaws in AI systems. The failures uncovered by Evals are concrete and actionable, making them amenable to mitigation through additional training, fine-tuning, or reinforcement learning from human feedback (RLHF) \citep{christiano2017deep}. However, a major limitation of Evals is that failing to identify flaws does not indicate that a system is safe for deployment (absence of evidence is not evidence of absence). Many evaluated systems  (often LLMs) are poorly understood black boxes and can be highly sensitive to small input variations. The ad-hoc and subjective nature of Evals evaluations risks not rigorously accounting for these subtle changes, limiting their reliability as comprehensive safety assessments. 

\subsection{Construct-Oriented {\School}}
The Construct-Oriented {\School} leverages system responses to quantitatively measure underlying ``{constructs}'' that describe the system's behaviour at an abstract level. These constructs are typically based on existing theories of cognitive traits or capabilities, but they can also be solely inferred from system responses (e.g., using factor analysis). To measure these \textit{latent} constructs, this {\school} often employs a \textit{functional} link with the observed system behaviour and \textit{designed} tasks carefully controlling for confounding factors, possibly adapting tasks from the cognitive sciences literature.
This {\school}  primarily aims to gain a better \textit{understanding} of self-contained \textit{systems}, particularly their \textit{performance}.
The works in this {\school} are mostly authored by \textit{researchers} with a background in \textit{psychology/cognitive science}.
Archetypal examples are Guinet~\etal~\shortcite{GauthierAutomated2024}, which applies psychometric methods to infer the ability of LLMs to solve 8th-grade mathematics tests, and Momennejad~\etal~\shortcite{momennejad2024evaluating}, which proposes a protocol for evaluating cognitive capabilities in LLMs, operationalising these capabilities through variations in specific tasks to systematically investigate how these variations influence LLM responses.

The Construct-Oriented {\school} has the  advantage of providing measurements  that can be  robust to variations in the test set and that  can be more readily transferred from evaluation settings to real-world deployment scenarios. However, developing evaluation instruments within this {\school} is considerably challenging: it requires strong domain expertise and detailed mathematical modelling of cognitive phenomena. A limitation of this approach is its reliance on existing human psychology theories which may not always provide suitable accounts for  complex or  capabilities and traits of general-purpose systems.

\subsection{Exploratory {\School}}
\looseness=-1
The Exploratory {\School} begins by forming a hypothesis---often inspired by anecdotal observations---about a system's behaviour. Similarly to how psychologists test hypothesis for human and animal cognition, a set of tasks capturing  key features---such as reasoning steps, memory requirements or generalisation patterns---is \textit{designed} to systematically isolate the phenomenon in different scenarios and exclude alternative explanations.
The findings arising from testing the \textit{system} on those tasks, mostly in terms of \textit{observed} \textit{performance} and occasionally \textit{manual inspection} of \textit{behavioural features}, are combined (and, in some cases, compared to humans) to provide evidence supporting the considered hypothesis qualitatively describing a system's ``cognitive processes''. 
The exploratory approach contrasts with the Construct-Oriented {\school}, which usually relies on pre-established theoretical frameworks that are complemented by the quantitative measurements of constructs. The aim to \textit{understand} the AI system's cognitive processes is what chiefly distinguishes this {\school} from Benchmarking (which also focuses on  observed behaviour, but for the purpose of comparing systems). Here, the AI system's results on tests are only important insofar as they support or refute the hypothesis of interest. Work in this {\school} is chiefly conducted by \textit{researchers} in \textit{AI} or \textit{psychology/cognitive science} and they have been mostly applied to LLMs; archetypal works include The Reversal Curse \citep{berglund2024the} and MeltingPot \citep{agapiou2022melting}.

The Exploratory {\school} offers the ability to propose and test  hypotheses and provide glimpses into AI systems' inner mechanisms at  the representational level of analysis \citep{Marr:1982:VCI:1095712}. By examining behavioural traces, this {\school} can provide novel insights into cognitive processes that may not be captured by aggregate metrics alone. This {\school} also has limitations: first, making well-designed tests requires substantial expertise and effort; moreover, each work in this {\school} typically relies on a few bespoke tests to explore narrow hypotheses, making it challenging to synthesise the findings into a comprehensive account of an AI system's overall behaviour and tendencies. This, together with the focus on observed indicators---rather than developing models of latent constructs---can limit generalisability, making it difficult to draw broader conclusions about a system's properties.

\subsection{Real-World Impact {\School}}

While most evaluation paradigms seek to assess specific properties of AI systems, the Real-World Impact (RWI) {\School} measures the impact of AI systems when deployed in the real world. 
This {\school} leverages techniques from the social and clinical sciences by running (randomised controlled) trials where, most often, assistance by an AI \textit{system} to humans is considered as an ``intervention'' whose effect must be quantified, in terms of a change in \textit{aggregate} \textit{performance} on the considered task relative to the system's \textit{cost}. The goal is therefore \textit{comparing} different AI systems, or AI-assisted humans to humans alone.
Evaluations are generally carried out \textit{interactively} or on a \textit{fixed} dataset and, due to the complexity of real-world tasks, human \textit{subjective} ratings or \textit{rubrics} may be employed. A common motivation for these evaluations is the \textit{comparison} of systems situated in the context of real-world applications requiring \textit{generation} of novel outputs. An archetypal work from this {\school} is \texttt{Math Converse} \citep{collins2023evaluatinglanguagemodelsmathematics}, which investigates the perceived helpfulness of LLMs for mathematics. A second representative work is Si~\etal~\shortcite{Si2024CanLG}'s study of LLM's ability to generate novel research ideas; which observes that human reviewers find LLM-generated ideas to be more novel but less feasible than those crafted by humans.

The RWI {\school} has a number of advantages: as AI systems become better at complex tasks, the social sciences provide many established methodologies to evaluate their societal impact, which may not be well estimated in artificial scenarios not  directly considering user experience. This {\school} has practical challenges as well: in contrast to other {\school}s, conducting experiments with human participants in realistic scenarios adds ethical constraints and logistical complexities. Moreover, this type of research mostly operates on a slower timescale compared to AI, making it hard for the RWI {\school} to timely provide information on new systems. 
These challenges have likely contributed to RWI being a small {\school} so far (it is the least represented in our sample of papers).  Nevertheless, we expect this {\school} to grow significantly in the coming years as AI systems become more capable of performing economically valuable tasks.

\subsection{ TEVV {\School}}
\looseness=-1
The Test, Evaluation, Verification, and Validation (TEVV) {\School} draws on methodologies from formal software verification, with its primary focus on ensuring that AI \textit{systems} behave in a well-defined and predictable way. TEVV is characterised by focusing on \textit{observed} \textit{extreme} values of \textit{safety} or \textit{robustness and reliability} measures, with the central goal of \textit{assurance}---providing bounds or guarantees for a minimum or average level of performance under various conditions. To achieve this, TEVV often explicitly operationalises the constructs it aims to measure, formally defining them to reduce uncertainty of the measurement process. It then employs either \textit{designed} tasks or \textit{operational} studies, with a variety of protocols. 
Works in this {\school} commonly deal with Reinforcement Learning and applied fields such as autonomous driving; 
Yang~\etal~\shortcite{Yang2023Safe} and Mussot~\etal~\shortcite{mussot2024assurance} are archetypal examples of TEVV works in these two fields.
 
TEVV offers several advantages, the most notable being its ability to provide formal guarantees and robust safety assurances. However, this approach requires a deep understanding of the system and its operational mechanisms. For many state-of-the-art or general-purpose AI systems, such an understanding is often lacking, making TEVV challenging to apply effectively. Indeed, we found that TEVV was one of the least represented {\school}s in recent AI venues.

\subsection{Evaluations crossing {\School} Boundaries}
\label{sec:crossing}
\looseness=-1
Unsurprisingly, we found several evaluation papers do not fit neatly into a single {\school}, but instead bridge multiple {\school}s and combine their methodologies (see also Fig.~\ref{fig:main}). For example, Perez~\etal~\shortcite{perez2022discoveringlanguagemodelbehaviors}'s model-written evaluations straddle the Benchmarking and Evals {\school}s, using LLMs to generate a large and varied set of questions, thus creating a standardised benchmark, that aim to provide safety assurances in the style typical of Evals. Similarly, CogBench \citep{pmlr-v235-coda-forno24a} bridges the Capability-oriented and Exploratory {\school}s. Here, the authors build a cognitive phenotype of LLMs using psychological experiments designed to assess different \textit{constructs}. Simultaneously, they \textit{explore} a number of hypotheses based on anecdotal evidence, including whether RLHF makes LLMs more human-like and the relationship between model size and tendency to exhibit human-like behaviour. These hybrid approaches demonstrate the flexibility of AI evaluation methodologies. 

\section{The Role of the Different {\School}s}
\label{sec:findings}
\looseness=-1
Each of the {\school}s we described plays a distinct role within the AI evaluation ecosystem. Each targets a particular type of measurement and fulfils the unique needs of different evaluators. For example, a company developing a safety-critical AI system, such as an autonomous vehicle, would rely on  evaluation methodologies that closely follow paradigm to obtain robust guarantees. On the other hand, a deployer of a system in a lower-stakes environment, such as an image classifier for pets, would likely rely on the performance of the system on a benchmark of representative images to determine when the system is ready to deploy. Techniques from multiple {\school}s can (and should) be combined when appropriate. This is already common among developers of state-of-the-art general-purpose AI systems, where a mix of Benchmarking (to assess capabilities such as reasoning or coding skills) and Evals (to assess safety concerns) is used. Beyond this specific combination, integrating methodologies across different {\school}s remains relatively uncommon. While this may be considered as a limitation of the current AI evaluation ecosystem, this also presents an opportunity for improvement.

\section{Challenges and Opportunities}
\label{sec:new_works}

\looseness=-1
Some {\school}s, such as Benchmarking, are applied across a wide range of AI systems. However, others tend to be domain-specific: TEVV is mostly applied to embodied, agentic forms of AI, such as RL systems or self-driving cars; similarly, in the papers we surveyed, we found the Evals and Exploratory {\school} were almost exclusively applied to LLMs, although this could be partly due to our focus on recent works. We believe this has occurred due to the way these {\school}s emerged in response to specific developments in AI. For example, Evals largely developed as a response to risks from LLMs \citep{ganguli2022red}. 
This means that a vast range of existing evaluation approaches remain underutilised in various domains and AI system types. While technically challenging, expanding the applications of different {\school}s beyond their typical uses would lead to a more comprehensive understanding of AI systems, their strengths, weaknesses, and broader impacts. While there appears to be growing interest in expanding the range of techniques applied to LLMs, often drawing on methods from TEVV \citep{huang2024verification,openai2024modelspec}, we hope to see this cross-pollination across all domains where AI is evaluated. 

Besides expanding the domain of application of each {\school}, great opportunities lie in developing new evaluations bridging different {\school}s, as the examples mentioned in Sec.~\ref{sec:crossing}. By highlighting the possibilities afforded by the {\school}s we identified, we hope to inspire researchers to develop new evaluations leveraging the strengths of multiple {\school}s for specific questions, to achieve more comprehensive and insightful assessments. At the same time, as discussed in Sec.~\ref{sec:findings}, using multiple paradigms to tackle an individual question from different perspectives is also a powerful but underexploited strategy.

We can integrate our insights with existing discussions on gaps in AI evaluation. 
For instance, Hutchinson~\etal~\shortcite{hutchinson2022evaluation} point out the lack of \textit{moral} evaluations in AI development raising important questions about data consent, the dignity of data workers, and the social responsibilities of developers. These factors were not included in our dimensional analysis due to the widespread lack of reporting on such topics. 
Similarly, Rauh~\etal~\shortcite{rauh2024gaps} identify a ``risk coverage gap'' with many ethical and social risks currently insufficiently addressed.  In our framework, evaluations addressing these risks would likely fall under the Real-World Impact {\School}, which we found to be the least developed. Therefore, Rauh~\etal~\shortcite{rauh2024gaps} and our work both highlight how this niche is unaddressed. 
Finally, Huang~\etal~\shortcite{huang2024verification} points out the lack of verifications with provable guarantees for LLMs, which also surfaced from our analysis of the TEVV {\school}. 
In general, by drawing attention to gaps in the space, we hope to encourage researchers to develop new evaluation methodologies that better address  issues.

\section{Limitations} \label{sec:Lims}
\looseness=-1
We aimed to capture the breadth of the AI evaluation landscape by surveying a highly diverse set of works. 
Given the extent of the field, our broad scope inevitably limited the depth with which we survey each {\school}. A focused investigation of fewer {\school}s might reveal additional patterns or relationships that we could not fully explore. Additionally, our selection of papers was based on our analysis of the recent literature, which may introduce bias. Certain evaluation paradigms, such as Benchmarking, may be overrepresented due to trends in the field, particularly the focus on LLMs. To mitigate this, we ensured that the authors of this paper have extensive AI evaluation expertise and deep familiarity with different areas of the ecosystem. This, combined with  seeking out evaluation works intentionally different from one another,  mitigated our selection bias as much as reasonably possible.

Another constraint lies in how our dimensions capture the nuances of different evaluation techniques. For example, we found the distinction between aggregate and functional distribution summary to be sometimes blurry---there is a fine line between stratifying aggregate performance based on predefined categories (e.g., required capabilities for tasks) and an imprecise functional model. This and other ambiguities can lead to disagreements among raters when assigning dimensions to papers and classifying them into particular {\school}s. At the same time, incorporating additional dimensions could offer deeper insights into how different {\school}s are characterised and help identify sub-{\school}s. Our approach balances granularity of annotation with practical usability.  
We further found that some dimensions were less informative than anticipated, for example Task Mode (Identification and Generation) was not useful for distinguishing between {\school}s where perhaps a more nuanced breakdown of Task Mode would have been.
Finally, the AI evaluation landscape is rapidly evolving and new {\school}s may emerge or existing ones become more refined. Despite these limitations, we believe the dimensions introduced here provide a valuable foundation for  guiding further research in AI evaluation and characterising future works.

\section{Conclusion}
\label{sec:conclusion}

This survey presents a snapshot of the current AI evaluation landscape, offering insights into prevailing approaches. We categorised over 125 recent or highly influential AI evaluation papers based on our multi-dimensional framework examining goals, methodologies, and research cultures. Through this analysis, we identified six distinct {\school}s offering individual perspectives to AI evaluation that contributes to the wider AI evaluation ecosystem, despite the lack of standardisation and occasional inconsistencies in terminology across these {\school}s. Our aim with this paper was to bring attention to these different approaches and foster greater cross-pollination between {\school}s, ultimately promoting a more integrated and holistic assessment of AI.

\newpage

\section*{Acknowledgements}
\looseness=-1
JB received funding from the Effective Ventures Foundation Long Term Future Fund through Grant ID: a3rAJ000000017iYA.
LP received funding from US DARPA (grant HR00112120007, RECoG-AI) and Open Philanthropy. JHO acknowledges the grants SFERA (PID2021-122830OB-C42) and Cátedra ENIA-UPV, TSI-100930-2023-9 funded by Spain's MCIN.

\bibliographystyle{named}
\bibliography{ijcai24} %

\end{document}